\documentclass[conference]{IEEEtran}
\IEEEoverridecommandlockouts
\usepackage{cite}
\usepackage{amsmath,amssymb,amsfonts}
\usepackage{algorithmic}
\usepackage{graphicx}
\usepackage{textcomp}
\usepackage{xcolor}
\def\BibTeX{{\rm B\kern-.05em{\sc i\kern-.025em b}\kern-.08em
    T\kern-.1667em\lower.7ex\hbox{E}\kern-.125emX}}
\begin{document}

\title{Deep CSI Learning for Gait Biometric Sensing and Recognition\\
\thanks{This work is supported through the INL Laboratory Directed Research \& Development (LDRD) Program under DOE Idaho Operations Office Contract DE-AC07-05ID14517}
}

\author{\IEEEauthorblockN{Kalvik Jakkala}
\IEEEauthorblockA{\textit{Department of Computer Science} \\
\textit{University of North Carolina at Charlotte}\\
Charlotte, USA \\
kjakkala@uncc.edu}
\and
\IEEEauthorblockN{Arupjyoti Bhuyan}
\IEEEauthorblockA{\textit{Idaho National Laboratory} \\
Idaho Falls, USA \\
arupjyoti.bhuyan@inl.gov}
\and
\IEEEauthorblockN{Zhi Sun}
\IEEEauthorblockA{\textit{Department of Electrical Engineering} \\
\textit{University at Buffalo}\\
Buffalo, USA \\
zhisun@buffalo.edu}
\and
\and
\hspace{2.5cm}
\IEEEauthorblockN{Pu Wang}
\IEEEauthorblockA{\hspace{2.5cm}\textit{Department of Computer Science} \\
	\hspace{2.5cm}\textit{University of North Carolina at Charlotte}\\
	\hspace{2.5cm}Charlotte, USA \\
	\hspace{2.5cm}Pu.Wang@uncc.edu}
\and
\IEEEauthorblockN{Zhuo Cheng}
\IEEEauthorblockA{\textit{Department of Computer Science} \\
	\textit{University of North Carolina at Charlotte}\\
	Charlotte, USA \\
	zcheng5@uncc.edu}
}

\maketitle

\begin{abstract}
Gait is a person's natural walking style and a complex biological process that is unique to each person. Recently, the channel state information (CSI) of WiFi devices have been exploited to capture human gait biometrics for user identification. However, the performance of existing CSI-based gait identification systems is far from satisfactory. They can only achieve limited identification accuracy (maximum $93\%$) only for a very small group of people (i.e., between 2 to 10). To address such challenge, an end-to-end deep CSI learning system is developed, which exploits deep neural networks to automatically learn the salient gait features in CSI data that are discriminative enough to distinguish different people  Firstly, the raw CSI data are sanitized through window-based denoising, mean centering and normalization. The sanitized data is then passed to a residual deep convolutional neural network (DCNN), which automatically extracts the hierarchical features of gait-signatures embedded in the CSI data. Finally, a softmax classifier utilizes the extracted features to make the final prediction about the identity of the user. In a typical indoor environment, a top-1 accuracy of $97.12 \pm 1.13\%$ is achieved for a dataset of 30 people. 


\end{abstract}


\section{Introduction}
Gait is a person's natural walking pattern. Medical studies have shown that gait is a very complex biological process and is unique to each person \cite{gait_intro}. There are studies that also show that when we try to imitate someone else's gait our own gait works against us\cite{gait_1,gait_2,gait_3}. These unique characteristics of gait make it ideal for user identification and authentication. Recently, the omnipresent WiFi devices have been used to capture human gait for user  recognition and authentication \cite{gait_CSI1, gait_CSI2,gait_CSI3}. When people walk through WiFi signals, their torso, legs, and arms cause unique variations. These variations contain the person's gait and can be captured via CSI. Gait can be captured with cameras or internal measurement units (IMU) \cite{imu_gait}. Unlike vision based methods, CSI-based gait identification relies on WiFi signals that can easily penetrate walls and obstructions, therefore working well in the cluttered environments full of obstructions. Different from the methods using IMU or wearable sensors, CSI-based gait identification is non-intrusive and device-free. Lastly, with the flourishing Internet of Things (IoT) deployment, omnipresent WiFi devices enable a ubiquitous and invisible security system 
through CSI-based biometric sensing. 

\begin{figure}[h]
	\centering
	\includegraphics[width=\columnwidth, keepaspectratio]{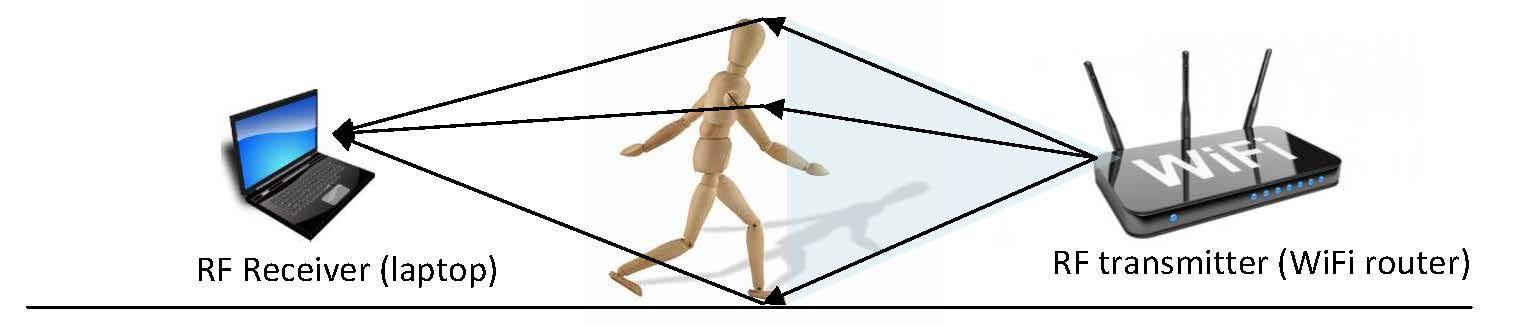}
	\vspace{-20pt}
	\caption{User identification through gait biometrics embedded in WiFi signals}
	\label{fig:room}
\end{figure}

Despite the aforementioned advantages of using CSI for gait sensing and identification, existing CSI based  identification systems have limited recognition accuracies (top-1 93\%) and are limited to a small number of people (2-10). The limiting factors in existing research are their reliance on hand-coded features such as minimum, maximum, kurtosis, the standard deviation of CSI data in the time domain and, entropy, energy and spectrogram signatures of the CSI data in frequency domain. Apart from hand-coded features, another major limiting factor in existing research is the use of rudimentary machine learning methods such as multilayer perceptrons and support vector machines. These fundamental limitations reduce the accuracy of their approach as the number of people in their dataset increases. 

\begin{figure}[h]
	\centering
	\includegraphics[width= 0.6\columnwidth, keepaspectratio]{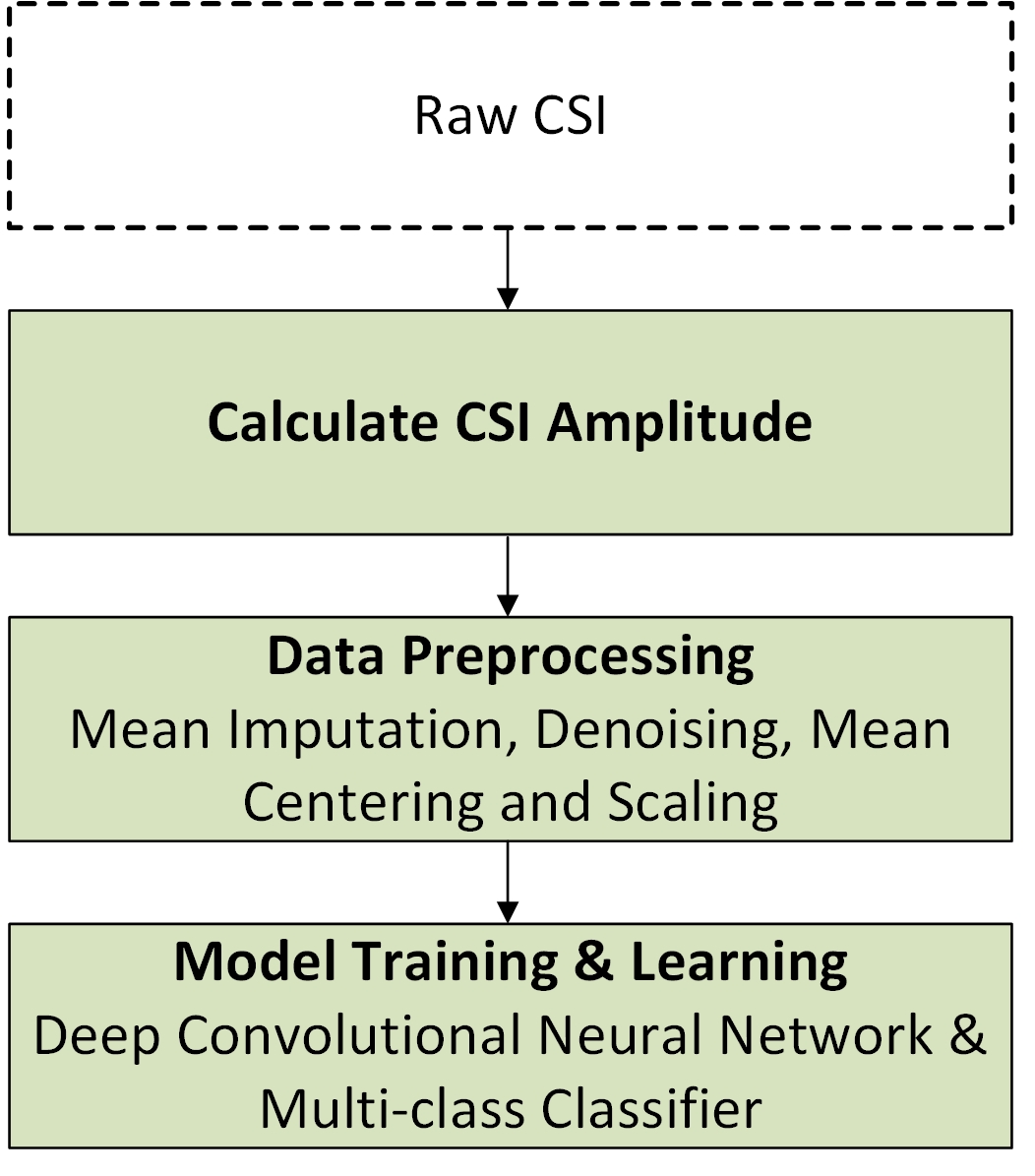}
	\caption{system Overview}
	\vspace{-10pt}
	\label{fig:overview}
\end{figure}

In this work, we propose a deep CSI learning system shown in Fig. \ref{fig:overview}, which addresses the aforementioned limitations of existing research.  Fundamentally different from the existing work, our system did not depend on any hand-crafted features. Instead, deep neural networks are used to automatically learn the salient gait features in CSI data. These features are discriminative enough to distinguish different people. In our system, the raw CSI data are first sanitized through window-based denoising, mean centering and normalization. The sanitized data is then passed to a residual deep convolutional neural network, which automatically extracts the hierarchical features of gait-signatures embedded in the CSI data. Finally, a softmax classifier utilizes the extracted features to make the final prediction about the identity of the user.  Our system achieves a top-1 accuracy of 97\% on a dataset consisting of 30 people in an indoor environment. This paper greatly improves the performance of our prior work on the gait recognition through deep CSI learning \cite{neuralwave}, which only achieves around $87\%$ identification accuracy for 30 people. The major limitations of our previous work are that it relied on PCA for lossy dimension reduction, which may unintentionally remove some useful and essential features that greatly affect the identification accuracy. Moreover, in this work, we adopt a different neural network architecture, which turns out to be more effective for gait feature extraction from CSI sensing samples.
%

The remaining sections of this paper are organized as follows: Section II introduces the overall system design, Section III presents the data preprocessing steps followed by Section IV which introduces our feature extraction and classification model. Lastly, in Section V, we show the experimental validation results and conclude the paper in Section VI.

\section{System Design}
In this section, we introduce some preliminaries of CSI followed by the data, then present the system challenges, and finally provide an overview of the overall framework. 

\begin{figure}[h]
	\centering
	\includegraphics[scale=0.9]{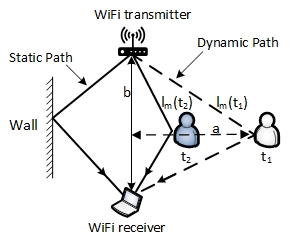}
	\caption{Human movement induces the length change of the paths reflected
		by the human body. The static environment objects, e.g., walls
		and furnitures, lead to signal reflections with unchanged paths.}
	\label{fig:dyn}
\end{figure}

\subsection{preliminaries}
Due to multi path effect, the CSI can be represented as follows:
\begin{equation}\label{cfr}
H(f;t)=  \underbrace{\sum_{n \in {p^s}} a_n \, e^{-j2\pi f \tau_n}}_{H_{static}(f)}  + \underbrace{\sum_{m \in p^d} a_m(t) \, \xi(f) \, e^{-j2\pi f \tau_m(t)}}_{H_{dynamic}(f;t)}
\end{equation}
Where $p^s$ is the set of static paths whose length is fixed, $p^d$ is the set of dynamic paths whose lengths keep changing due to moving objects. As illustrated in Fig.~\ref{fig:dyn}, for one dynamic transmission path $m\in{p^d}$, its length will change from $l_m(t_1)$ to $l_m(t_2)$. $a_n$ and $a_m$ are the attenuation of each path in $p_s$ and $p_d$ separately. $\xi(f)$ is the frequency dependent absorption cross section coefficient (ACS) caused by the body signal absorption \cite{melia2013electromagnetic}. $\tau_m(t)$ is the propagation delay for path $m\in{p^d}$.

The Doppler frequency shift is contained in dynamic part of CSI. When a person walks, the length of path $m$ changes, the signal reflected by the moving human buddy via path $m\in{p^d}$ has Doppler frequency shift:

\begin{equation}\label{fdm}
f^D_m(t)=\frac{1}{\lambda}\frac{dl_m(t)}{dt}=f\frac{d\tau_m(t)}{dt}
\end{equation}

Thus, $H_{dynamic}(f,t)$ can be rewritten as:

\begin{equation}\label{doppler}
H_{dynamic}(f;t) = \sum_{m \in p^d} a_m(t) \, \xi(f) \, e^{-j2\pi  \int_{-\infty}^t f_m^D(u) du}
\end{equation}

From \eqref{doppler} we know that CSI contains doppler frequency shift information. Knowing the environment setting shown in Fig.~\ref{fig:dyn}, we derive the Doppler frequency shift theoretically, from \eqref{doppler}. In our system, a person starts walking from a distance of $a$, from the transmitter and receiver, at a speed of $v$. The distance between the transmitter and receiver is $b$. We only consider the subject's torso which causes high energy macro Doppler shifts and ignore the low energy micro shifts from the subject's limbs.\textsf{} By calculating the propagation path length change, the theoretical Doppler frequency shift would be:

\begin{equation}\label{fmd}
f^D_m(t) = \frac{(a-vt)(-v)}{\sqrt{(b^2/4)+(a-vt)^2}}/\lambda
\end{equation}

\begin{figure}[h]
	\centering
	\includegraphics[scale=0.14]{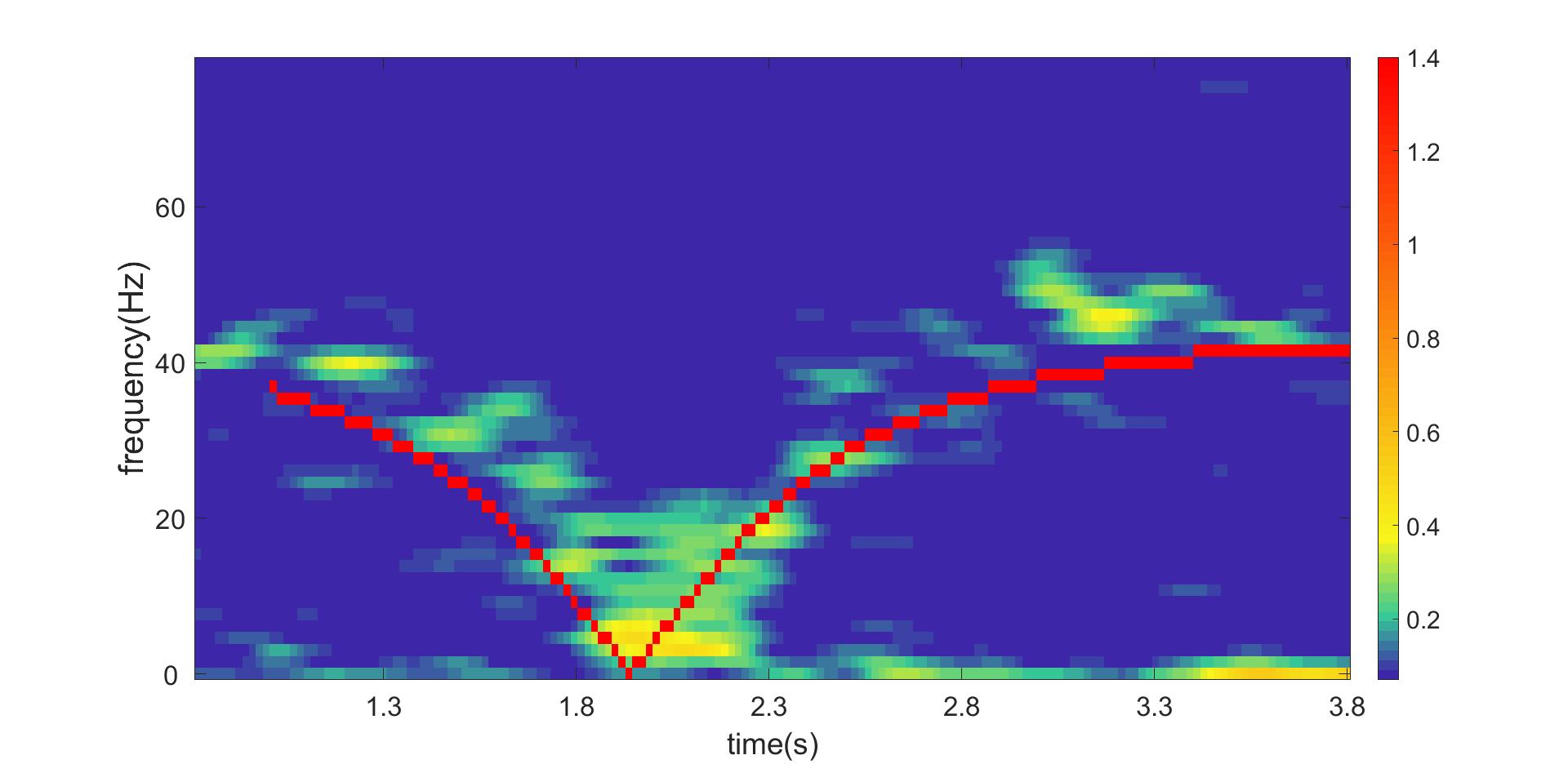}
	\caption{Spectrogram of a CSI data sample. The red line is the theoretically derived Doppler frequency shift.}
	\label{fig:dyn_der}
\end{figure}

We compute Fast Fourier Transformation (FFT) on a CSI data sample and plot the spectrogram to observe whether it matches our theoretical derivation in \eqref{fmd}. As shown in Fig.~\ref{fig:dyn_der}, the spectrogram matches our theoretical derivation, showing that the CSI data does contain a user's gait signature.

\subsection{CSI Data Sample}
The CSI data was collected from a pair of off-the-shelf 802.11n WiFi transmitter and receiver with  ($N_{rx}$ = 3) receiver antennas and ($N_{tx}$ = 3) transmitter antennas operating on ($N_{C}$ = 30) subcarriers. The CSI measurements between a pair of transmitting and receiving antennas over one subcarrier during a measurement period $T$ will constitute one CSI waveform. Therefore, there exist 270 CSI waveforms for our system setup. All the 270 CSI waveforms in turn form a CSI sample. Since each packet will generate one CSI measurement, with a transmitting data rate of 2000 packets/second and measurement period $T = 4$ second, each CSI sample can contain $270 \times 8000$ features.

\subsection{Challenges}
One of the major challenges faced by deep CSI learning is the noisy and high dimensional CSI data samples. Our prior work addressed this issue with a wavelet filter for noise suppression and the principal component analysis (PCA) for dimension reduction. Even though PCA works well for dimension reduction that eases the training of followed deep neural network, it loses any structural and non-linearly correlated information. Therefore, our prior work was only able to get a top-1 accuracy of 87.76\%. To address this challenge, we first resample the raw CSI streams with much lower frequency and then let deep neural network perform feature dimension reduction and feature learning.



\section{Data Preprocessing}
The CSI data collected from off-the-shelf WiFi devices are noisy and incomplete due to the interference-rich environments, hardware imperfections, and internal state transitions on both the transmitter and receiver. Moreover, CSI data contains unnecessary static components, which are introduced by the signal reflections from the static obstructions, such as walls and furnitures. To obtain clean CSI data, data preprocessing has to be performed. 

\subsection{Mean Imputation}
\begin{figure}[h]
	\centering
	\includegraphics[width=\columnwidth, keepaspectratio]{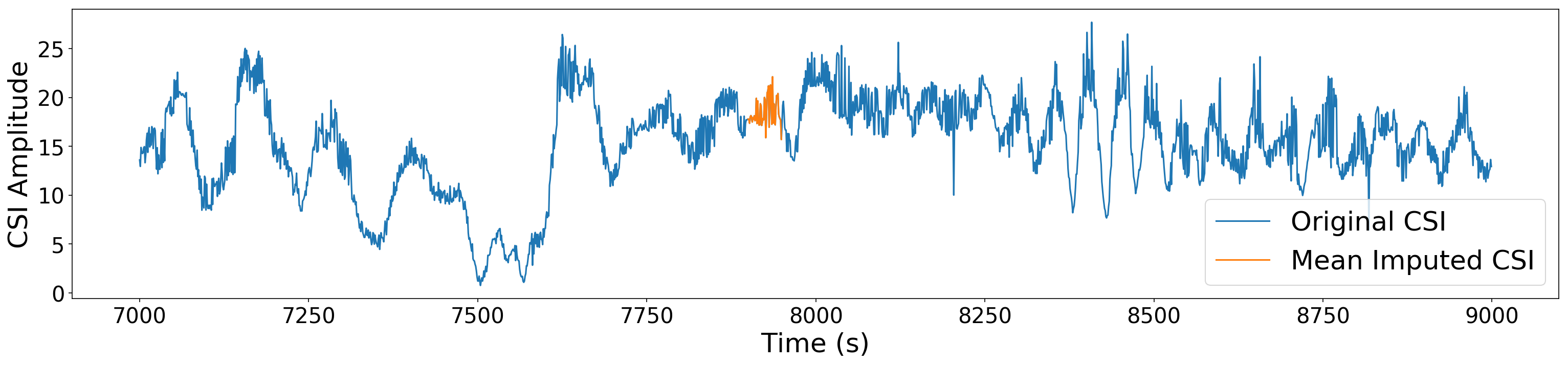}
	\caption{Mean imputed waveform}
	\vspace{-10pt}
	\label{fig:impute}
\end{figure}

WiFi devices dynamically select the number modulation and coding set (MCS) of the connection between the transmitter and receiver. This changes the number of transmitting antennas and therefore the number of CSI waveforms we receive in each CSI packet varies over the period of data collection. There are several operations defined in literature to deal with missing data such as regression imputation and K nearest neighbors \cite{imputation}. We address this issue with mean imputation. We pad the missing waveforms with the mean of the waveforms we receive at that time step. Since all the waveforms we receive in a single time step are highly correlated, this operation does not change the distribution of the data and allows us to maintain a consistent data dimension for every sample. We apply this operation to data samples in both training and testing sets and, the operation is not computationally expensive, allowing us to apply the operation in real time implementations as well. Fig.~\ref{fig:impute} shows an example of mean imputation for a CSI packet with missing waveforms.  

\subsection{Denoising and Resampling}
\begin{figure}[h]
	\centering
	\includegraphics[width=\columnwidth, keepaspectratio]{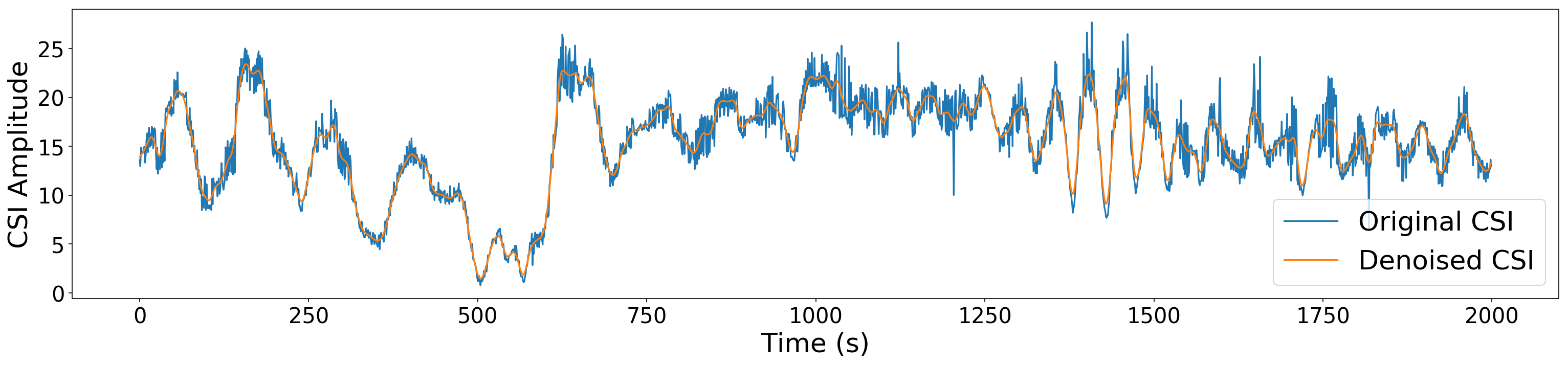}
	\caption{Original waveform vs denoised waveform}
	\vspace{-10pt}
	\label{fig:denoise}
\end{figure}

We use a simple window function to reduce the noise in our raw CSI samples. In particular, we use a Hanning window of length $N_{w}$ = 91. The window function convolves a Hanning window with a predefined length across each column of the data sample. Similar to mean imputation the window function is computationally efficient and good for real-time implementations. Moreover, the window function is good for filtering out high-frequency noise while preserving the underlying structure of the waveform as shown in Fig.~\ref{fig:denoise}. After denoising, we re-sample each CSI data sample with a dimension of $270 \times 8000$ using a sampling rate of 125$Hz$. This makes the  CSI sample have a reduced dimension of $270 \times 500$.

.

\subsection{Mean Centering and Scaling}
Before we can use a DCNN for the final classification, the data needs to be mean-centered and scaled. We split our dataset into training and testing set ie., $X_{train}$ and $X_{test}$. The mean, minimum and maximum of the training set alone are computed which are then used for mean centering and scaling both training and testing data. We compute only the training set's statistics to avoid data leakage, i. e., training a machine learning model with data outside of the training set. Avoiding data leakage allows us to make sure the model generalized to unseen real-world data. Lastly, mean centering apart from decreasing training time and improving model performance, also removes the static components of the CSI data which, contains irrelevant information about the data collection environment, such as the walls and furniture in a room.

\section{Deep CSI Learning}
\subsection{Motivations for Deep CSI Leaning}
Unlike prior work, we rely on deep learning for feature extraction, compared to traditional hand-coded techniques. With deep learning, we can extract high-level features relevant for user recognition while ignoring features irrelevant for user recognition, such as the environmental details. Deep neural networks are stacks of operations designed to extract high-level features from the input data. DCNNs are a  special class of neural networks which contain convolutional layers. A convolutional layer convolves banks of learnable weights across a given input \cite{DeepLearning}. This operation is ideal for extracting spatially correlated information. Usually, the output of a convolutional operation is passed to a batch normalization (BN) layer followed by an activation layer. BN layers regularize the data by maintaining the mean and variance of batches of training data and, uses them to normalize the data. This operation reduces the risk of overfitting i,e., when the model memorizes the training set instead of generalizing to the unseen real world data. While the activation function introduces non-linearities in the model \cite{DeepLearning}, individual convolutional layers are limited to linear feature extraction. But, by stacking these operations and introducing activation functions the model can extract nonlinear features. DCNNs are well suited for extracting gait information from CSI data as it contains a lot of spatial correlations in its time domain form. Since we rely on gait information whose features are unlikely to appear in the same time window, convolutional layers being shift-invariant ie., insensitive to time shift of features of interest in the data, are an ideal solution for such feature extraction. 

\begin{figure}[h]
	\centering
	\includegraphics[scale=0.5]{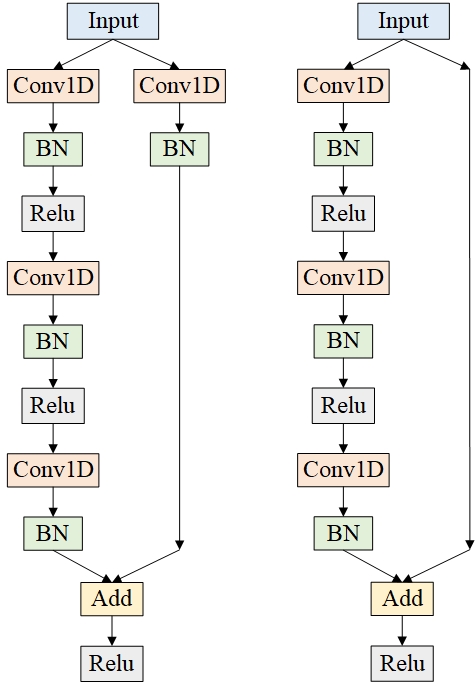}
	\caption{Convolutional residual block (left) and Identity residual block(right)}
	\vspace{-10pt}
	\label{fig:conv_res_new}
\end{figure}

\begin{figure}[h]
	\centering
	\includegraphics[width=\columnwidth, keepaspectratio]{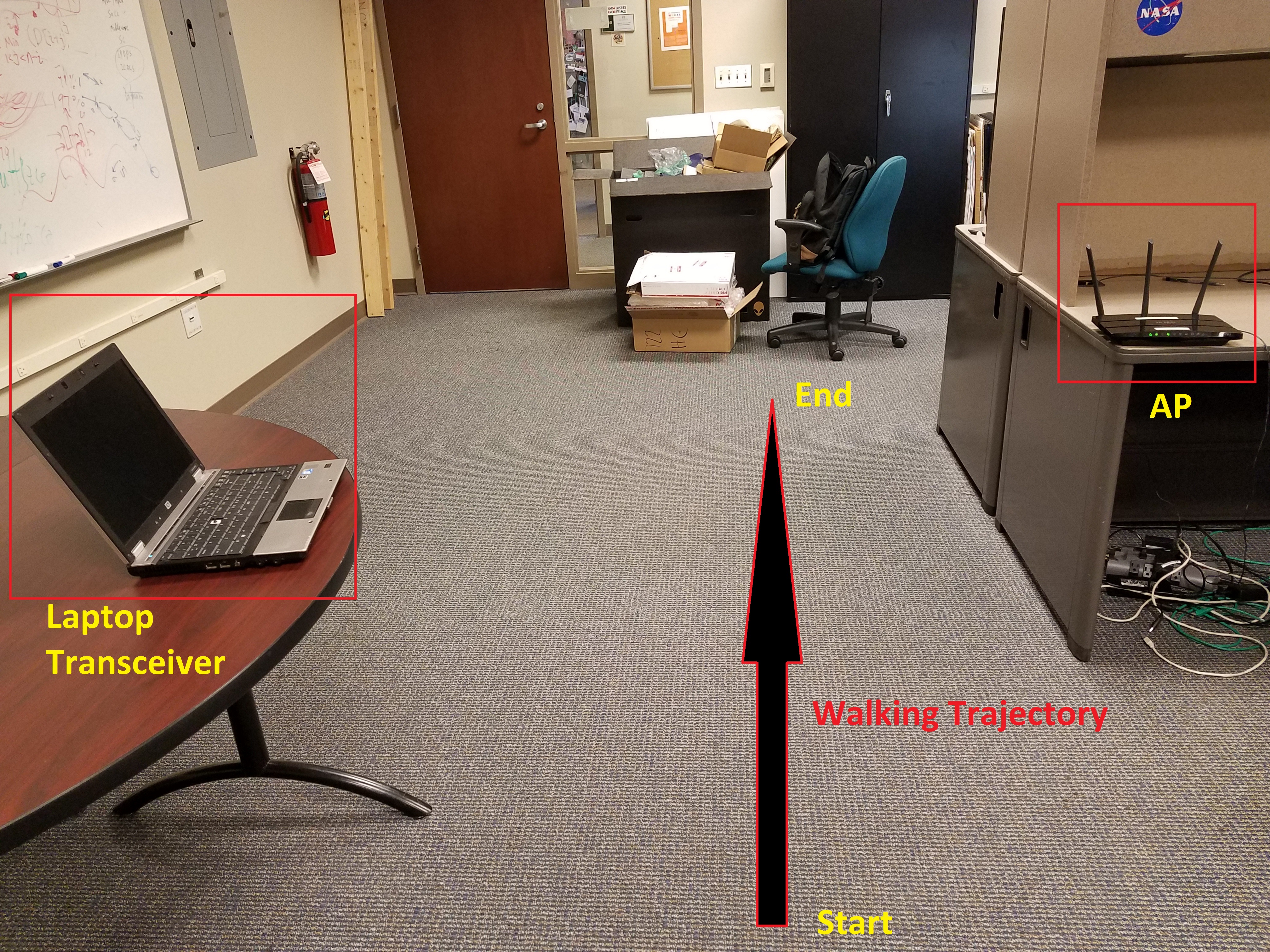}
	\caption{Data collection environment}
	\vspace{-10pt}
	\label{fig:room_dat}
\end{figure}

\subsection{Model Architecture}
There are several neural network architectures in the literature which use convolutional layers. After applying several key architectures, we settle in a 18 layer Resnet proposed in \cite{ResNet}. The distinguishing characteristic of Resnets is that they contain residual blocks. Residual blocks contain groups of convolutional, activation and BN layers and, are organized in a way that they map the input data of the block to each succeeding layer within the block and also have skip connections to subsequent layers. This allows for better gradient flow during backpropagation in the training phase of the model. The model we used contains two types of residual blocks, identity and convolutional. Identity residual block map the input data to the subsequent block unimpeded while convolutional residual blocks have a convolutional layer in the skip connections as shown in Fig.~\ref{fig:conv_res_new}. The overall architecture is made up of stacked residual blocks followed by a multinomial softmax classifier \cite{DeepLearning}. The softmax classifier uses the final extracted features to make a prediction about the identity of the user. 

\subsection{Dataset and Training Methodology}
The data collection environment is shown in Fig. \ref{fig:room_dat}, where we placed a WiFi router which transmitted with a data rate of 2000 packets per second to a laptop with an Intel 5300 NIC. The receiving laptop calculates CSI data for every packet received from the transmitter. We use the CSI extraction tool introduced in \cite{Halperin_csitool}. The CSI measurement from each data packet contains up to 270 raw features. We consider $4$-second CSI measurements to form a CSI sample. We collected 1303 CSI data samples from 30 volunteers with their age ranging from 21 - 40. The dataset was split into a training and testing sets with an 85-15\% split respectively. The resent was trained using the Adam optimizer \cite{DeepLearning} with 1e-3 learning rate and 1e-2 decay rate for 30 epochs. 

\section{Experimental Validation Results}
This section introduces some of the evaluation metrics used to quantify the artificial neural network performance, analyze the results and mention some techniques that did not work.

\subsection{Accuracy}
Training and testing accuracies are tow fundamental evaluation metrics used in machine learning. DCNNs are trained on a subset of a dataset called the training set. The accuracy for the training set accuracy is the percentage of correct prediction for samples in the training set.

\begin{equation}
Train \ Accuracy = \frac{\text{Total number of correct predictions}}{\text{Total number of train samples}} \nonumber
\end{equation}

\begin{equation}
Test \ Accuracy = \frac{\text{Total number of correct predictions}}{\text{Total number of test samples}} \nonumber
\end{equation} 

Similar to train prediction, testing accuracy is the percentage of correct predictions from the test set. But, unlike the training set, the testing set is not used to update the weights of DCNN. This allows us to quantify the real world performance of the model. Our framework achieves $97.12 \pm 1.13\%$ mean testing accuracy. Fig.~\ref{fig:confu} shows the average training and testing accuracy of our framework after training the model on 20 randomly selected train and test sets.

\begin{figure}[h]
	\centering
	\includegraphics[width=\columnwidth, keepaspectratio]{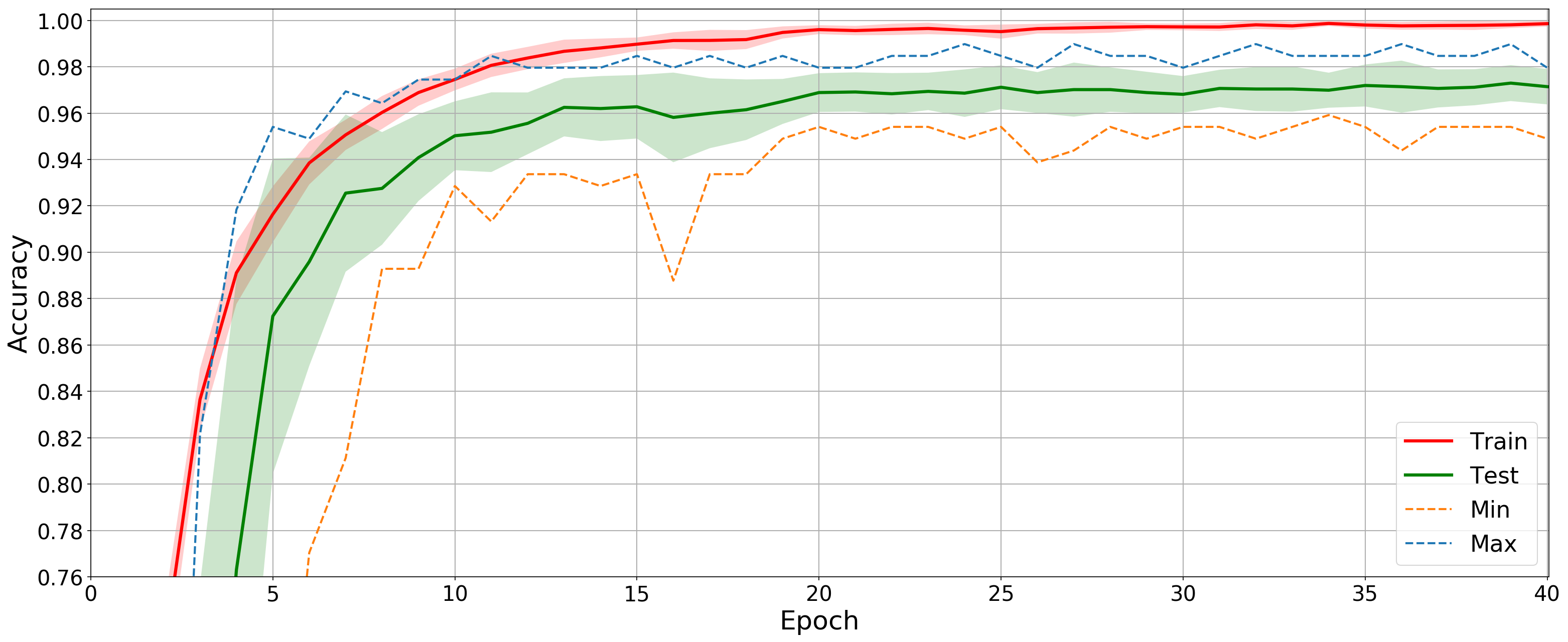}
	\caption{Train and test accuracy of 20, 18-layer resnets each trained on a randomly selected train and test set.}
	\vspace{-10pt}
	\label{fig:acc}
\end{figure}

\subsection{Confusion Matrix}
\begin{figure}[h]
	\centering
	\includegraphics[scale=0.21]{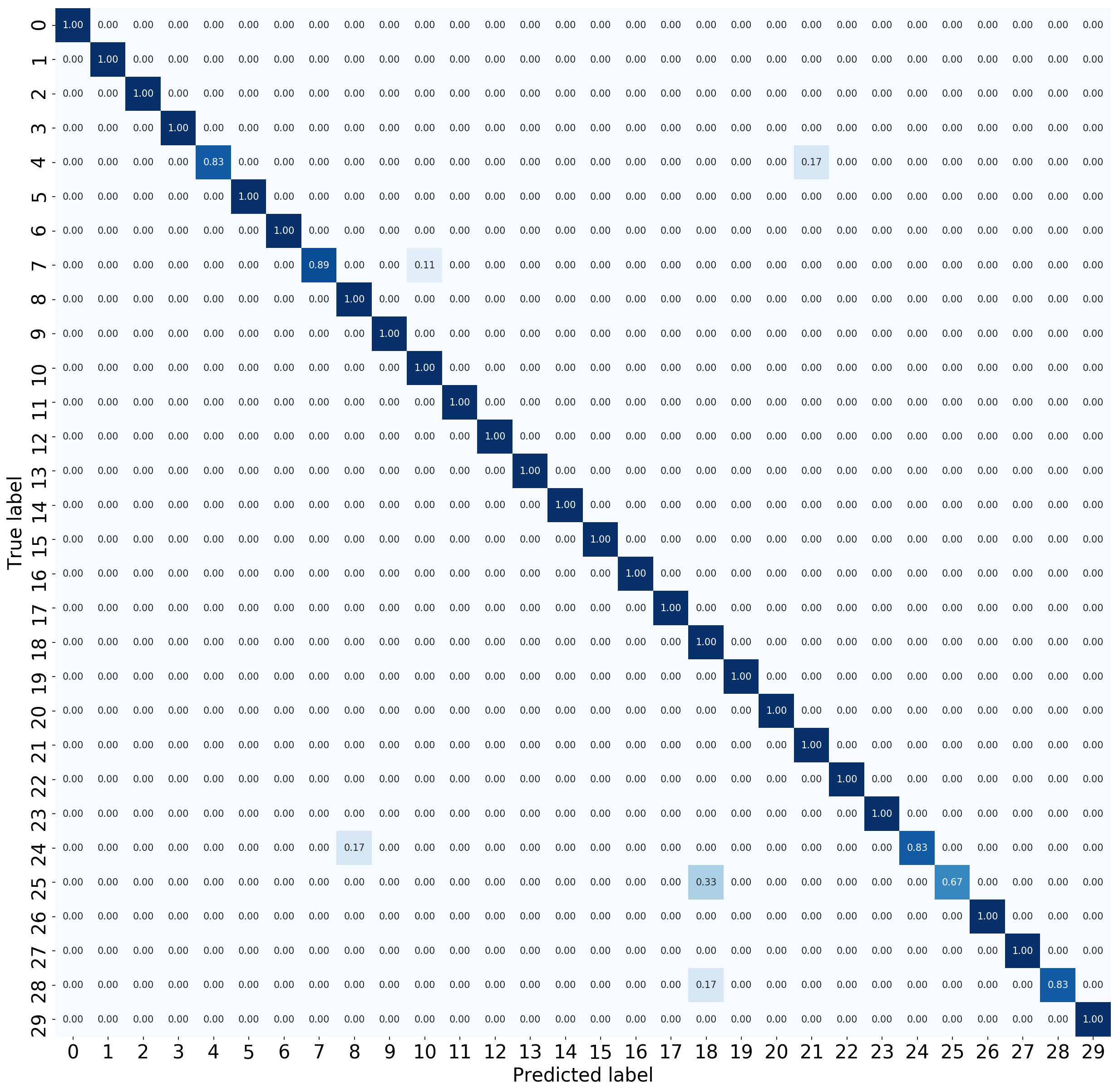}
	\caption{Confusion matrix generated from resnet that achieved $96.94\%$ identification accuracy}
	\vspace{-10pt}
	\label{fig:confu}
\end{figure}

The confusion matrix is used to get better insight into how good the model is at classifying data from a specific class. The confusion matrix contains four metrics true positive (TP), true negative (TN), false positive (FP) and lastly false negative (FN). A sample is positive if it belongs to a given class and negative when it does not belong to that class. Based on this terminology TP value is the numbers of samples which are positive and get a positive prediction, TN is the numbers of samples which are positive and get a negative prediction, FP is the numbers of samples which are negative and get a positive prediction and lastly, FN is the numbers of samples which are negative and get a negative prediction. Fig.~\ref{fig:confu} shows the confusion matrix for our framework, its performance is consistent across all classes, which all have high TP along with low FP and FN.

\begin{figure}[h]
	\centering
	\includegraphics[scale=0.4]{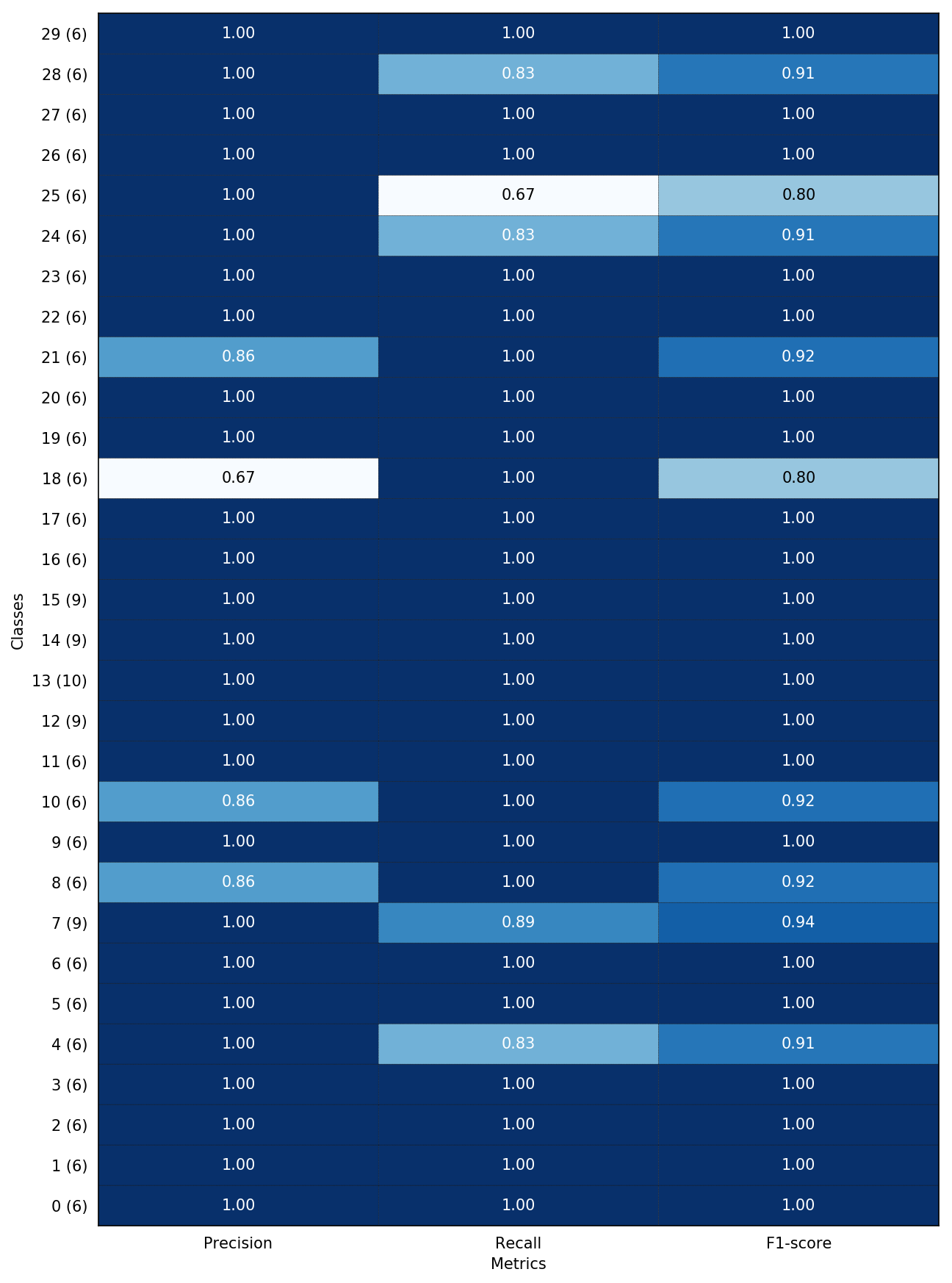}
	\caption{Classification report generated from resnet that achieved $96.94\%$ identification accuracy}
	\vspace{-10pt}
	\label{fig:classi_rep}
\end{figure}

\subsection{Classification Report}
Another popular evaluation metric is the classification report which actually a combination of 3 metrics, precision, recall, and F1 score. The three metrics are computed from values in the confusion matrix ie., TP, TN, FP, and FN. Precision is the confidence of the classifier to identify relevant samples of a class. The recall is the quantification of the ability of the classifier to identify all relevant cases in the dataset. Lastly, the F1 score can be interpreted as the weighted average of precision and recall. Fig.~\ref{fig:classi_rep} shows the classification report of our system, which, shows that the majority of the classes have high precision, recall and F1 scores. 

\begin{equation}
\centering
\begin{split}
\small
Precision = \frac{TP}{(TP+FP)} \,\,\,\,\,\,\, Recall = \frac{TP}{TP+FN} \\ F1_{Score} = \frac{2(Recall \times Precision)}{(Recall + Precision)}
\end{split}
\end{equation}  

\subsection{Frequency Domain Classification}
Apart from classifying time-domain data, we also tried frequency domain classification. We generated spectrograms from the first PCA component of every data sample and found the data to be very sparse and easy to overfit. We achieved a top-1 accuracy of 60\% on the test set with spectrogram images using an 18 layer resnet. We also tried generating 5 spectrograms from a single data sample by taking the first 5 PCA components. These spectrograms were then stacked to form a 3-dimensional tensor which was considered as a single data sample. We obtained a limited  60\% test accuracy with an 18 layer resnet when trained with the 3-dimensional spectrogram tensors. The possible reasons why frequency-domain (spectrogram) learning is not effective are as follows. Each CSI sample contains 270 CSI waveforms, which correspond to 270 spectrograms that forms a 270-D image as a single input tensor for the neural network. To reduce the dimension of this tensor, we need reduce the number of the spectrograms generated by each CSI sample. Thus, we convert the CSI sample into PCA domain and use the first PCA-CSI component to create a single spectrogram as  a 2D image tensor. However, such spectrogram can only show the aggregated Doppler frequency shifts from 270 CSI waveforms, thus losing fine-grained details for user identification. 

\section{Conclusion}
In this paper, we proposed an end-to-end deep CSI learning system to extract and identify gait signatures from WiFi signals. Our system sanitizes and resamples the raw CSI samples to generate low-dimensional clean CSI samples. Such data samples are used to train a 18-layer residual DCNN, which automatically learns the salient gait signatures embedded in the CSI samples. These learned  features are distinctive enough to enable accurate user identification. In particular, our proposed system is able to achieve a top-1 test accuracy of $97.12 \pm 1.13\%$ on a dataset of 30 human subject.


\bibliographystyle{IEEEtran}
\bibliography{refs,gait}

\end{document}